# The influence of motion features in temporal perception

**Rosa Illán Castillo[1]**, **Javier Valenzuela[2]**

[1]Laboratoire Dynamique Du Langage (Lyon) – CNRS
maria-del-rosario.illan-castillo@cnrs.fr
[2]Universidad de Murcia
jvalen@um.es

**Abstract**

This paper examines the role of manner-of-motion verbs in shaping subjective temporal perception and emotional resonance. Through four complementary studies, we explore how these verbs influence the conceptualization of time, examining their use in literal and metaphorical (temporal) contexts. Our findings reveal that faster verbs (e.g., *fly*, *zoom*) evoke dynamic and engaging temporal experiences, often linked to positive emotions and greater agency. In contrast, slower verbs (e.g., *crawl*, *drag*) convey passivity, monotony, and negative emotions, reflecting tedious or constrained experiences of time. These effects are amplified in metaphorical contexts, where manner verbs encode emotional and experiential nuances that transcend their literal meanings. We also find that participants prefer manner verbs over path verbs (e.g., *go*, *pass*) in emotionally charged temporal contexts, as manner verbs capture the experiential and emotional qualities of time more effectively. These findings highlight the interplay between language, motion, and emotion in shaping temporal perception, offering insights into how linguistic framing influences subjective experiences of time.

## 1. Introduction: Time and motion

Over the past four decades, scholars have converged on the idea that humans conceptualize time primarily in terms of space (TIME IS SPACE). The connections between the domain of time and the domain of space have been known since antiquity (e.g., Guyau, 1890) and have been developed throughout the decades (e.g., Clark, 1973; Traugott, 1978). In the 1980s and 1990s, conceptual metaphor theorists provided abundant evidence of the systematic spatial construal of time in language, arguing that it was actually a matter of thought, not merely surface linguistic expression. In fact, the study of time spatialization has flourished in both cognitive linguistics and psycholinguistics as a test bed for the grounding of abstract thought in sensorimotor experience (Lakoff & Johnson, 1980; Lakoff, 1993; Boroditsky, 2000; Casasanto & Boroditsky, 2008; Ishihara et al., 2008; Ulrich & Maienborn, 2010). This intersection between time and motion offers a fertile ground for exploring how linguistic structures reflect and even shape human experience.

TIME IS SPACE constitutes a metaphor which is thought to be experientially motivated, since our experience of motion is clearly correlated with our experience of time (Grady, 1997). Nonetheless, as stated by Evans (2004), these mappings between time and space cannot be considered primary metaphors, due to the complexities that time reveals at different levels: subjective experience, cultural and cross-linguistic differences or mapping gaps, among others. Furthermore, even though these two concepts are experientially close, time and space differ in many ways. For instance, while space is three-dimensional and isotropic —uniform in all its directions—, time is one-dimensional and has an asymmetric organization —it can only flow in one direction: it moves forward, never backwards—. We can remember the past and plan for the future, but we cannot change or revisit the past. Time events in motion are all aligned, with fixed relative positions: Monday and Tuesday do not arrive simultaneously or undertake one another. These examples evince that spatiotemporal metaphors expressing motion constitute complex models of temporality rather than relatively simple sets of cross-domain mappings.

Our preconceptual experience of time, which is influenced by our sensory experiences and the nature of the external world, contributes to our understanding of time in complex and nuanced ways. Time is not a single, uniform phenomenon, but rather a diverse range of events and processes that occur at different levels of experience. Objective time and perceived time are different constructs. Although the "speed" of time remains constant —as C.S. Lewis said, *The future is something which everyone reaches at the rate of sixty minutes an hour*— our personal experience of time is a highly subjective phenomenon. The perception of duration is influenced by a wide range of external and internal factors —such as affective states (James, 1890; Fraser, 1978)— that vary from person to person. The nuanced experience of time perception is a profoundly intricate subject, touching upon a universal aspect of human existence that has captured the interest of scholars from various fields since the 1800s, including philosophers, psychologists, neuroscientists, and linguists, among others (e.g., Friedman, 1990; Macey, 1994; Varela, 1999; Rao et al., 2001; van Wassenhove et al., 2008; Andersen & Grush, 2009; Eagleman, 2008; Bar-Haim et al., 2010; Falk, 2013; Evans, 2013). This phenomenon has been explored through diverse lenses, revealing multiple factors that influence how we perceive time. Emotions, for example, are widely recognized to significantly impact our sense of time, with fear or impatience seemingly stretching moments (illustrated by the saying *A watched pot never boils*), while joyous moments appear to pass more swiftly (*Time flies when you are having fun*), as noted by Agarwal &

Karahanna, 2000. Age also plays a crucial role in shaping our temporal perception, with older individuals often feeling that time passes more quickly than when they were young. The impatient query *Are we there yet?* often posed by children during travel, epitomizes the variation in temporal perception between younger individuals and adults, highlighting the subjective nature of experiencing time (see Droit-Volet & Meck, 2007 for a review).

Time and its associated concepts (duration, sequence, aging, biography, perceptions of temporality and their related emotions) require substantial cultural work to be built, a considerable cognitive development to be learned and adequate representations to be transmitted. How, then, are these complexities expressed through language? Which linguistic strategies do speakers use to convey the subjectivity of temporal perception and its flow? Corpus-based studies of sentences including a temporal unit functioning as the subject of a motion verb (Valenzuela & Illán Castillo, 2022; Illán Castillo, 2024) have shown that, while motion verbs encoding path seem, *prima facie,* to be the most common way for English and Spanish speakers to refer to the simple, linear progression of time (e.g. *The days go by*, *Christmas is coming*), speakers clearly prefer choosing a manner of motion verb when they want to encode specific meanings associated with their subjective experience of the passing of time (e.g. *The days crawl by*, *Time is limping along*). This fact is not so obvious: one might initially assume that path verbs would be the most useful type for temporal metaphor usage. Commonly referenced mappings in scholarly works (e.g. Lakoff & Johnson, 1980, 1999; Núñez & Sweetser, 2006) draw parallels between the components of physical movement and those intervening in a temporal scene. For example, the Source of the motion corresponds to the beginning of the temporal stretch, the Goal equates to its conclusion; the distance traveled represents the time passed, and the distance remaining matches the time left before the period concludes, among other analogies (Valenzuela & Illán Castillo, 2022):

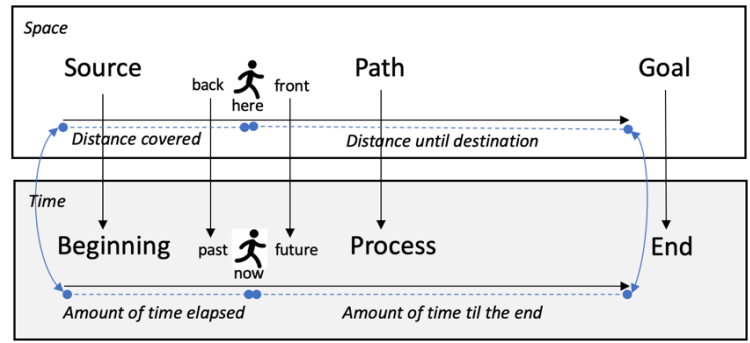

Figure 1. Some mappings between SPACE and TIME (in an Ego-Moving perspective)
(Valenzuela & Illán Castillo, 2022, p. 21).

Upon examining these conceptual mappings, the motion verbs that profile such elements (Source, Path, Goal, Distance, or Direction) fall into the category of path verbs (Talmy, 2000). This aligns with expectations, as these components are integral to what Talmy describes as the basic motion event (Talmy, 2000: 215). Yet, this basic structure is augmented by what is known as a Co-event, encompassing the Cause or the Manner of motion. These additional elements, though critical, are often overlooked in discussions of how motion and time are interconnected in conceptual mappings. After all, the role that the manner of motion could play in the realm of time is not immediately obvious.

At the heart of this study is thus an inquiry into the specific role that manner of motion verbs —verbs that describe the way in which movement occurs, such as *crawl* or *rush*— play in expressing temporal concepts. While previous research has explored how path verbs, which emphasize the trajectory or endpoint of movement, contribute to temporal metaphors, relatively little attention has been given to the role of manner verbs in this domain. This paper seeks to fill that gap by investigating how the manner of motion, particularly the speed of movement, influences the emotional and experiential nuances associated with the passage of time.

**2. Experimental studies**

In this section we present a series of four interconnected studies. These studies represent a closer look into the intricate dynamics of the emergent meanings that are generated through the use of motion verbs in temporal contexts. These studies are designed as exploratory tests, serving as a foundational approach to the nuanced analysis of how motion verbs, particularly those encoding manner, contribute to the conceptualization of time in language. The primary aim is to empirically validate and substantiate some of the observations and hypotheses formulated throughout previous corpus analyses of the conceptualization of time as motion (Valenzuela & Illán Castillo, 2022; Illán Castillo, 2024). By exploring these verbs in both metaphorical (temporal) and literal (physical) contexts, we aim to deepen our understanding of how manner features, interact with the conceptual domain of time to shape interpretation and emotional resonance.

Study 1 examines a range of features associated with manner of motion verbs, with a particular focus on speed. It explores how slow and fast verbs —as well as some other salient features— contribute to meaning construction and emotional interpretation when describing the passage of time.

Study 2 examines the global valence of manner of motion verbs in both literal (physical) and metaphorical (temporal) contexts. This study provides a comparative perspective on how these verbs are perceived emotionally when describing actual physical motion versus abstract temporal experiences.

In study 3, we delve deeper into valence, this time investigating how participants associate temporal expressions with emojis rather than numerical valence ratings. This approach highlights the relationship between metaphorical language and non-verbal emotional representations.

Finally, study 4 uses a fill-in-the-blank task with sentences containing explicitly encoded emotions to examine participants' preferences and associations. The study investigates how participants complete sentences with different types of manner of motion verbs (e.g., *crawl* vs. *fly*) or choose between manner of motion verbs and path verbs (e.g., *crawl* vs. *go*; *fly* vs. *pass*). This design aims to uncover patterns in how verb types interact with emotional contexts to shape interpretation.

Together, these studies provide a comprehensive examination of manner of motion verbs and their role in shaping temporal conceptualization and perception. By exploring how features like speed and motion dynamics interact with abstract temporal domains, we highlight the intricate ways language influences how we understand and emotionally experience time.

## 2.1. Study 1: Emergent meanings survey

**Participants**

This study involved 59 native English speakers, aged 22 to 54 years, who were recruited through Prolific. The participant group included 39 females and 20 males.

**Materials**

Participants were presented with 15 sentences —see Appendix I—, each featuring a motion verb with a Time Measurement Unit (TMU) as the subject. A Time Measurement Unit refers to a linguistic expression that quantifies time, such as *minutes*, *hours*, *days*, *weeks*, *months*, *years*, and *centuries*, all of which depict measurable stretches of time in human experience. These TMUs serve as temporal markers that, when combined with motion verbs, create metaphorical expressions of time's movement. The selected verbs for this study were *approach, fly, hasten, race, rush, slip, slide, spin, drag, creep, crawl, limp, edge [closer], inch [closer]*, and *run*. These verbs were chosen for their potential to convey varying perceptions of time's passage and its emotional and subjective nuances, as shown by previous corpus-based studies on the lexicalization patterns of spatiotemporal metaphors (Valenzuela & Illán Castillo, 2022; Illán Castillo, 2024). The nine questions designed to elicit participants' interpretations of the sentences were as follows:

1) How is the passage of time perceived? (Very slowly to Very quickly)
2) Time is passing in a (Monotonous, boring way to Lively, entertaining way)
3) How many things are happening? (Nothing is happening to Lots of events are happening)
4) Are events under control? (No control over the events to Events are fully controlled)
5) How productive is the speaker? (They achieve nothing to They get a lot of things done)
6) Is there a sense of imminence, something is about to happen? (No to Yes)
7) Is there a sense of surprise of the event happening? (No to Yes)
8) Is the speaker impatient of what is to come? (No to Yes)
9) Is time passing with difficulty, are there some setbacks during the day? (No to Yes)

These questions were crafted to encompass various dimensions of temporal perception and the emotional resonances evoked by the sentences, thereby facilitating a comprehensive analysis of the emergent meanings generated by motion verbs in temporal contexts.

**Procedure**

The survey was administered online via Qualtrics. Participants were instructed to read each of the 15 sentences carefully and then respond to a series of 9 questions regarding the temporal and emotional implications conveyed by the use of the motion verb in context. For each question, a slider ranging from 1 to 7 allowed participants to quantify their perceptions, with an additional option to select *Not apply* where appropriate. The questions aimed to capture a spectrum of temporal experiences, from the speed of time's passage to the emotional and situational context implied by the verb's use.

**Results and Discussion**

In the analysis of the responses from the 59 native English speakers to the provided 16 sentences featuring Time Measurement Units (TMUs) as subjects of motion verbs, several key trends emerged. As expected, across the 15 sentences, verbs indicative of faster motion (*fly*, *hasten*, *race*, *rush*, *slip*, *slide*, and *spin*) were consistently associated with perceptions of quicker time passage, as opposed to verbs suggesting slower movement (e.g., *drag*, *creep*, *crawl*, *limp*), which correlated with slower time perceptions. Speed emerged as the central manner feature activating further emotional meanings related to the speaker's subjective interpretation of time. The variation in speed, as conveyed by the different motion verbs, significantly influenced participants' perceptions of time's passage —ranging from 'slow and monotonous' to 'fast and lively/entertaining'—. This suggests that speed is not merely a descriptive attribute but a critical cognitive cue that shapes the emotional and experiential qualities attributed to temporal events. Participants generally rated sentences with motion verbs that imply a fast pace as indicating more lively and entertaining experiences of time passing. This suggests a link between speed and a more dynamic perception of time. Conversely, 'slow rate' manner of motion verbs clearly elicited perceptions of monotony.

The strong positive correlation observed between the perceived speed of motion verbs and levels of entertainment highlights the role of speed as a fundamental dimension through which individuals interpret and emotionally engage with time. Faster speeds were consistently associated with more dynamic and engaging temporal experiences, as indicated by higher entertainment ratings. Conversely, slower speeds were linked to perceptions of time passing tediously, highlighting speed's influence in modulating the emotional valence associated with temporal flow.

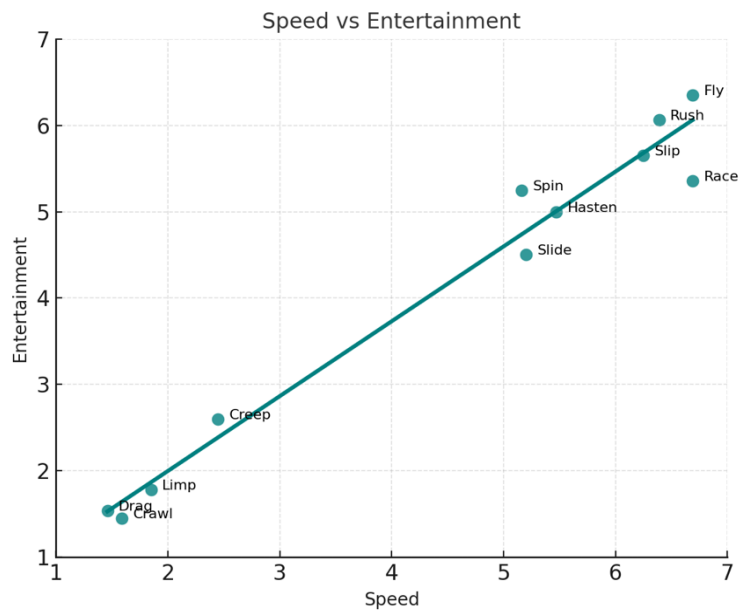

Figure 2. Pearson correlation between *Speed* and *Entertainment*

Participants' perceptions of entertainment and enjoyment showed a very strong positive correlation with speed ($r = 0.985$), as illustrated in Figure 1. The linear trend observed in the plot, fitted with a regression line, indicates that as the motion verbs conveyed greater speed, participants reported an increased sense of liveliness and engagement, underscoring the intrinsic link between the rapidity of motion and the intensity of entertainment experienced.

Another interesting link was found between speed and the participant's perception of the number of events occurring within a given timeframe. The statistical analysis revealed a very strong positive correlation between these two variables, with a correlation coefficient ($r = 0.963$). This finding suggests that the faster the action implied by the verb, the more events participants imagined as occurring. This relationship is depicted visually in the plot below:

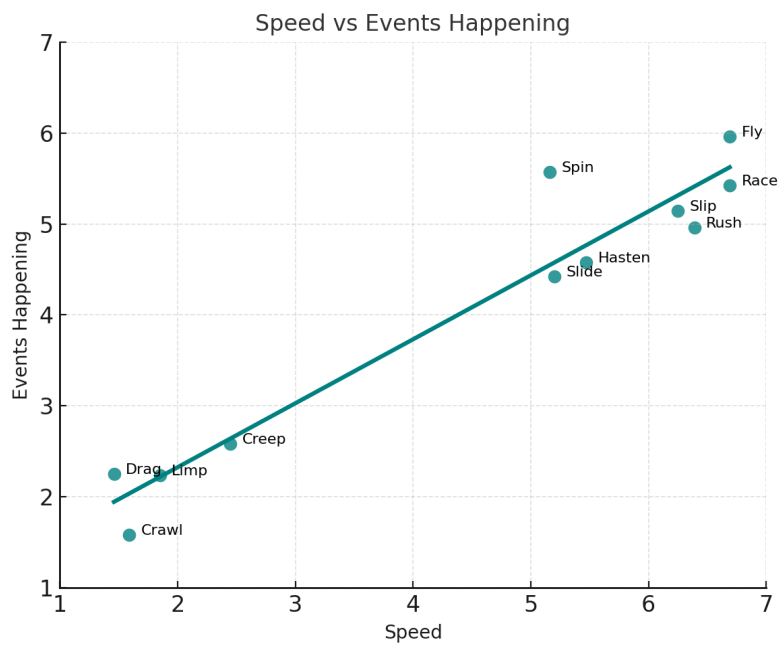

Figure 3. Pearson correlation between *Speed* and *Events Happening*

Figure 3 shows the correlation between speed and the perception of events happening. The upward trend in the data points, fitted with a regression line, illustrates the very strong positive correlation, where verbs that suggest faster motion correspond with narratives that are perceived as more eventful. According to this, the verbs that denote quicker actions (*fly*, *hasten*, *race*, *rush*, *slip*, *slide*, and *spin*) not only convey a physical quickness but also seem to carry with them a narrative density that is felt by the participants. These results align with the idea that, in situations where attention is diverted from the passage of time (like during an engaging activity), arousal may lead to a speeding up of perceived time (Izard, 2009). Therefore, putting it the other way around, if an event is perceived as having a rapid motion associated with it, speakers tend to link it to aspects like eventfulness and entertainment, even when they have no context to determine this, as is the case in the stimuli selected for this experiment. For example, a sentence like *The minutes hasten in this place* is not giving information about the number of events happening or aspects related to the level of entertainment. All this information is derived from the use of the verb *hasten* associated with time. This, once again, proves the huge capability of manner verbs to activate and generate meanings in temporal contexts that are not explicitly expressed or codified in the language.

Results also indicate that the speed conveyed by a motion verb affects how challenging the associated actions are perceived to be. Our findings show a strong negative correlation between speed and perceived difficulty ($r = -0.931$), presented visually in Figure 3. The downward trajectory depicted in the plot suggests that actions characterized by faster motion verbs were perceived as less difficult. This nuanced relationship may reflect the fluency and efficiency of time's passage, contrasting with the laborious and challenging nature often attributed to slower movements.

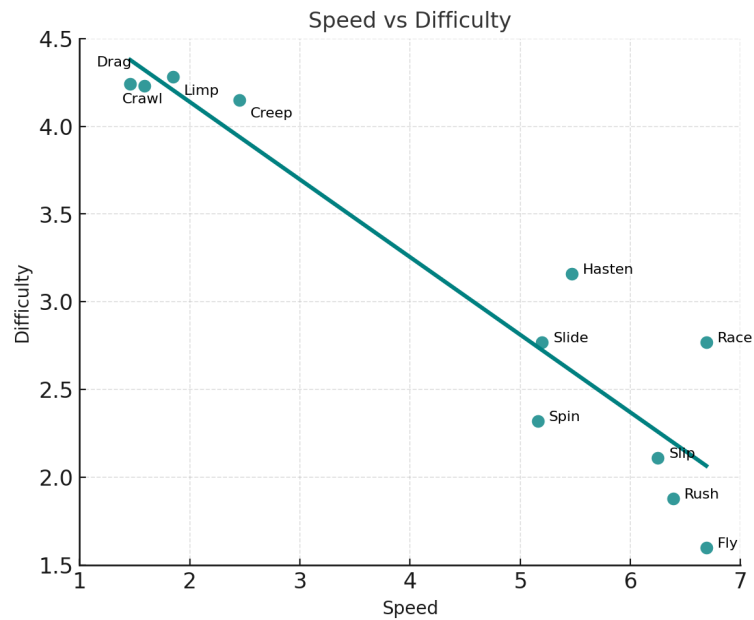

Figure 4. Pearson correlation between *Speed* and *Difficulty*

This phenomenon reveals a nuanced cognitive processing of temporal language, where rapidity does not equate to increased cognitive load but rather suggests a streamlined progression of events. It reinforces the idea we have been proposing: that our subjective experience of time, as manipulated by the speed of motion verbs, extends beyond mere pace, impacting our emotional engagement with the narrative and our perceived ability to navigate within it.

Continuing with speed, it is important to note that the verb *run* was not rated high in this variable. This confirms previous corpus-based observations (Valenzuela & Illán Castillo, 2022; Illán Castillo, 2024) that this verb's nuance of 'fast rate' is usually not activated when the verb is used in temporal contexts.

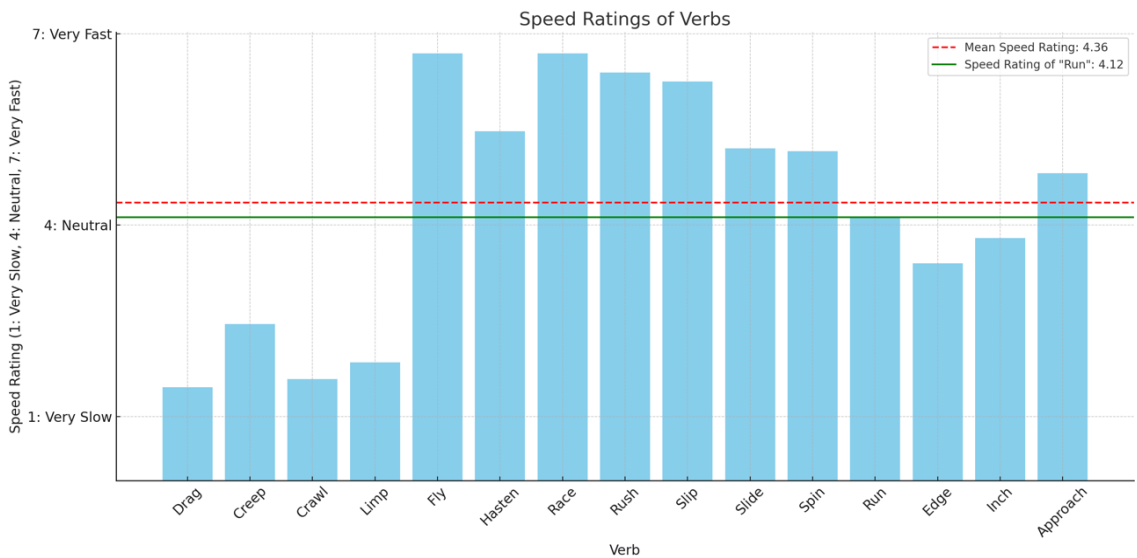

Figure 5. Speed rating means across verbs

In the visual representation of Figure 5, it is evident that the verb *run* garners a neutral rating, situated closely around the midpoint of the speed scale. It is important to note that *run* was also rated high in 'control' over the events happening (with a mean of 6.2 out of 7), since its use is usually linked to expressing information about a schedule or a specific point in time in which some event takes place, as seen in corpus examples. In addition, the plot also shows that *approach* receives a rating that leans towards the faster end of the spectrum. This can be attributed to its association with imminence, a quality that imbues a sense of urgency and thereby, a faster perception. Similarly, *edge* and *inch*, when used in conjunction with *closer*, manifest higher speed ratings than might be expected given their conventional association with gradual or slow movement. This outcome suggests that the context of nearing or closing in, regardless of the inherent motion speed indicated by these verbs, amplifies the perceived rate of action.

  The aforementioned verbs were found to trigger for the anticipation of something about to occur. While typically associated with proximity, the term *imminence* conveys not only closeness in time or space but also a heightened sense of urgency or potential peril. This broader interpretation aligns with definitions such as the one found in the *Cambridge Dictionary* ('[especially of something unpleasant] likely to happen very soon', n.d.) which exemplifies *imminent* with contexts of threat, danger, or negative anticipation, suggesting an event that is not just near, but also looming with significant impact. The survey responses indicate that these verbs carry with them a sense of tension, a forewarning.

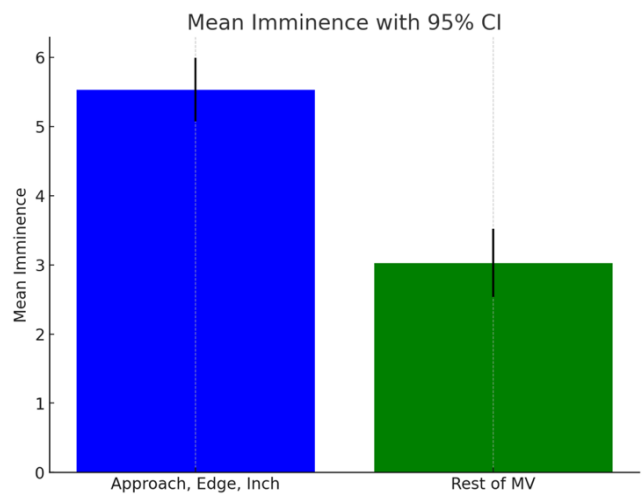

Figure 6. Comparative analysis of mean imminence across divergent verb groups

A t-test for independent samples yielded a t-statistic of 10.68 and a *p*-value of approximately 2.02e-07 ($p < 0.05$). This result suggests a statistically significant difference between the mean imminence scores of the verbs *approach*, *edge closer*, and *inch closer* and the rest of MVs, as depicted in Figure 6.

Another factor which seems to play a role is how much "control" the agent has over the events happening. The analysis of control ratings across verbs with varying implied speeds provides valuable insights into how participants perceive agency over time-related events. The overall mean control score across all verbs was found to be 3.88 on a scale from 1 (no control) to 7 (full control), suggesting a general perception of limited control. This below-neutral average score implies that while individuals may feel some level of agency, there is an underlying sense that the passage of time is often beyond personal control. However, further examination revealed that this sense of control varies meaningfully with the perceived speed of actions.

When categorizing verbs by speed, a clear distinction emerged between "fast" verbs (*fly*, *hasten*, *race*, *rush*, *slip*, *slide*, and *spin*) and "slow" verbs (*drag*, *creep*, *crawl* and *limp*). The mean control score for fast verbs was higher at 4.15 compared to the lower mean of 3.56 for slow verbs. This difference was statistically significant (t = 3.36, p < 0.001) and was further quantified with a Cohen's d effect size of 0.32, indicating a small to moderate effect. The higher ratings for fast verbs could be attributed to the dynamic, actively initiated nature of these actions, which likely fosters a sense of agency. In contrast, verbs describing slower movements, such as *creep* and *limp*, imply constrained or passive progression, where time seems to advance independently of one's will. A visual comparison further supports this finding:

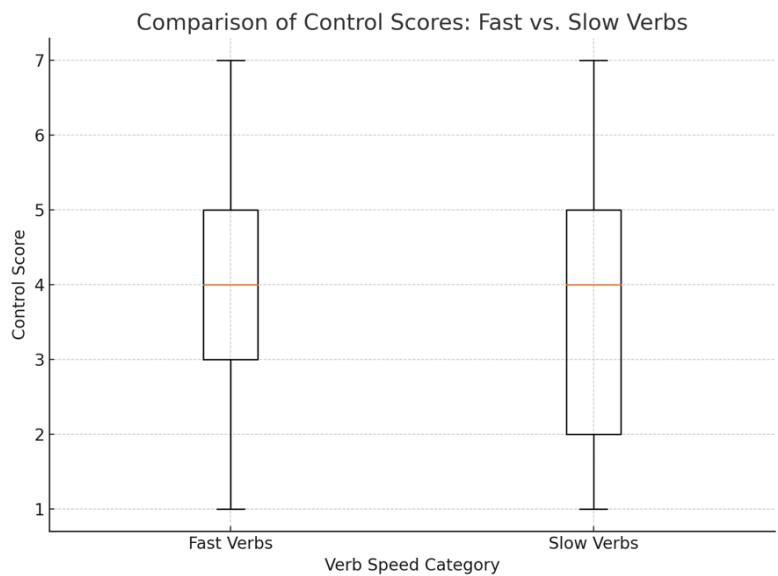

Figure 7. Comparison of *Control* scores

The box plot of control scores for fast and slow verbs shows that fast verbs have a higher median control score and a smaller interquartile range (IQR), indicating that participants' control ratings for these verbs are both higher and more consistent. This suggests that participants perceive a relatively stable level of control over faster actions, likely because these actions are associated with active, positive, intentional engagement. In contrast, slow verbs not only have a lower median control score but also a broader IQR. This wider spread indicates that perceptions of control over these slower actions vary more across participants. The broader range suggests that, while slower actions are generally perceived as less controllable, there is greater variability in these ratings—perhaps reflecting differing interpretations or situational factors that influence the perception of agency in slower, more constrained actions.

The association of faster motion verbs with greater perceived control and slower motion verbs with less control may initially seem counterintuitive. One might expect that with faster motion, there would be less control, as events moving quickly could feel more chaotic or harder to manage. Conversely, slower motion might intuitively suggest more control, as one might assume that a slower pace allows for more deliberate actions and choices. However, the metaphorical use of these verbs to describe the passage of time introduces different psychological dynamics.

In the context of temporal perception, faster motion verbs (e.g., *fly*, *rush*) are often linked with positive, engaging, and lively experiences where time subjectively feels accelerated. During these enjoyable or immersive events, people frequently report a greater sense of agency, feeling as though they are actively shaping their experience rather than passively enduring it. This engagement aligns with a higher perception of control, despite the quickened pace. Slower verbs (e.g., *drag*, *crawl*), on the other hand, tend to describe monotonous, negative, or tedious experiences where time feels sluggish, resulting in diminished engagement and agency. Here, the slow pace actually conveys a lack of control, as people often feel "stuck" in time, passively waiting for events to end. This pattern reveals how perceptions of agency over time may be shaped by the nature of activities, with faster, dynamic actions fostering a stronger sense of control and slower, constrained actions diminishing it.

**Conclusions**

Study 1 has aimed to provide a step forward to unravel the nuanced interpretations of motion verbs when applied within temporal contexts. By closely examining qualities traditionally linked to physical motion —specifically speed and proximity— we have tried to shed light on the complex process by which these attributes assume novel significances in the realm of temporal expression. This preliminary task highlights the versatile semantic landscape of motion verbs when used in temporal contexts, bridging the gap between physical movement and the abstract passage of time to generate new meanings. Our results suggest that speed is more than a descriptive attribute; it acts as a cognitive cue that shapes the emotional quality of temporal experiences and the perceived control over these experiences. Interestingly, while faster motion might intuitively suggest less control, we observed that faster verbs not only signaled quick and lively time but also suggested greater agency, eventfulness, and anticipation. This finding likely stems from the association of faster motion verbs with positive, engaging events, where participants feel actively involved and thus more in control. In contrast, slower verbs conveyed monotony and difficulty, contributing to a more passive and uneventful experience of time, where control was perceived as diminished. The slower pacing, linked with negative or tedious events, fostered a sense of passivity, as participants felt more constrained by time. This interplay between speed, control, and perception highlights the role of motion verbs in shaping subjective experiences of time. In temporal contexts, motion verbs go beyond mere descriptors, influencing how actively or passively individuals perceive themselves in relation to the passage of time. These insights deepen our understanding of the cognitive and emotional processes involved in temporal language, illustrating how linguistic framing can alter one's sense of agency in experiencing time.

## 2.2. Study 2: Constructing meaning and valence - Manner of motion verbs in literal (physical) and metaphorical (temporal) contexts

**Participants**

This study involved 46 native English speakers (23 females, 23 males), aged 21 to 45, recruited through Prolific.

**Materials**

The materials consisted of 20 sentences featuring 10 manner of motion verbs (*crawl, creep, drag, edge, fly, inch, limp, linger, blow, zoom*) used in both literal and metaphorical temporal contexts. Each verb appeared twice (once for each condition, literal and temporal). The sentences were designed to elicit valence ratings on a scale from 1 (very negative) to 9 (very positive).

**Procedure**

The study was conducted online using Qualtrics. Participants read each sentence carefully and rated its emotional tone using an interactive slider corresponding to the 1-to-9 valence scale. This slider allowed for precise responses and provided a visual cue for the rating process. The instructions emphasized that there were no right or wrong answers and that participants should base their ratings solely on their personal interpretation of each sentence. To prevent repetition effects, each participant viewed only one of the two conditions (literal or temporal) for each verb. Sentences were presented in random order to control for order effects.

**Results and discussion**

The analysis revealed significant differences in valence ratings between slow and fast verbs and between literal and metaphorical contexts. Mean valence ratings for slow verbs (*crawl, creep, drag, edge, inch, limp, linger*) were lower in metaphorical contexts (M = 3.55) compared to literal contexts (M = 5.13), indicating a more negative perception of time when these verbs were used metaphorically. In contrast, fast verbs (*fly, blow, zoom*) showed higher valence ratings in metaphorical contexts (M = 6.51) compared to literal contexts (M = 5.44), aligning with perceptions of accelerated or positive time movement.

Wilcoxon tests confirmed significant differences between literal and metaphorical contexts for both groups of verbs. For slow verbs, the Wilcoxon test yielded a W-Statistic of 2887.0 and a p-value of 0.00000423, indicating a significant decrease in valence ratings for metaphorical usage. For fast verbs, the Wilcoxon test produced a W-Statistic of 373.5 and a p-value of 0.000165, confirming a significant shift toward more positive valence in metaphorical contexts.

A two-way ANOVA was conducted to investigate the effects of verb group (slow vs. fast) and context (literal vs. metaphorical) on valence ratings, as well as their interaction. The analysis revealed a significant main effect of verb group, $F(1, 456) = 114.80$, $p < 0.0001$, indicating that fast verbs were rated more positively overall than slow verbs. Additionally, a significant main effect of context was observed, $F(1, 456) = 3.91$, $p = 0.0485$, suggesting that valence ratings differed between literal and metaphorical contexts. Crucially, there was a significant interaction effect between verb group and context, F(1, 456) = 34.39, p < 0.0001, demonstrating that the influence of context on valence ratings varied depending on the verb group. Specifically, slow verbs showed a decrease in valence in metaphorical contexts compared to literal contexts, while fast verbs exhibited the opposite pattern, with metaphorical contexts eliciting higher valence ratings

A grouped bar plot visually states these differences, showing a decline in valence for slow verbs in metaphorical contexts and an increase for fast verbs:

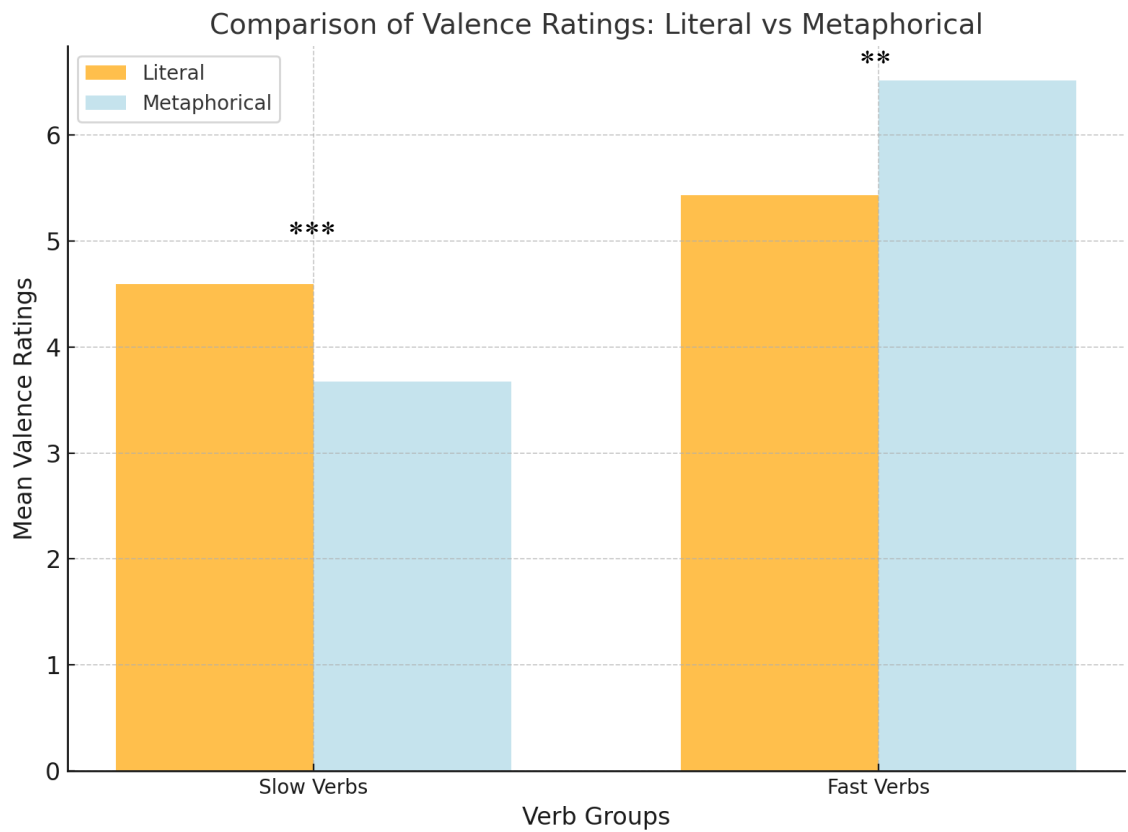

Figure 8. Comparison of valence ratings: Literal vs metaphorical

As shown in Figure 8, in their literal context, both slow and fast verbs cluster around a valence rating of 5, which corresponds to a neutral emotional tone. For instance, a sentence like *The ivy crept up* is interpreted as emotionally neutral, focusing on the physical act of movement without evoking strong emotional connotations. Similarly, fast motion verbs like *zoom* in a literal sense (e.g., *The motorcycle zoomed by*) do not carry a significant positive or negative emotional weight, maintaining their neutral stance.

However, in the metaphorical contexts where these verbs are used to describe the passage of time, their valence diverges significantly. For slow verbs, the metaphorical usage results in a substantial drop in valence ratings below 5. This indicates that phrases like *The minutes crept by* or *The day dragged on* evoke a more negative emotional tone. Such expressions often metaphorically map slow motion to undesirable experiences, such as boredom or frustration, reflecting a cognitive association between slow movement and an unpleasant passage of time or situation. In contrast, fast verbs in the metaphorical context show a significant increase in valence ratings above 5. Sentences like *Time flew by* or *The hours zoomed by* suggest a positive emotional experience, often linked to engaging or enjoyable events where time seems to move quickly.

This contrast highlights how the same verb can shift in emotional tone depending on its usage. In literal contexts, the verbs remain descriptive, focusing on physical motion without much emotional coloring. Yet, in metaphorical contexts, they acquire emergent emotional meanings tied to temporal experiences. Such findings invite reflection on the cognitive and linguistic mechanisms at play. Metaphors involving manner verbs not only describe time but also frame it in ways that influence our emotional response. The observed valence shifts reveal how metaphorical language actively contributes to our emotional and cognitive interpretations of everyday experience.

It is worth mentioning that, to minimize potential bias in valence ratings, we selected subjects with neutral or near-neutral emotional valences, as reported by Warriner et al. (2013). For literal contexts, subjects included words such as *turtle* (6.16), *plane* (5.72), *ant* (3.9), *snail* (4.52), *finger* (5.8), *motorcycle* (5.8), and *wind* (5.67). While these values are mostly close to neutral, they show a wider range, from *ant* (3.9, slightly negative) to *turtle* (6.16, slightly positive). For temporal contexts, subjects such as *minute* (5.5), *week* (5.27), *hour* (5.05), and *year* (5.75) exhibit even less variability, clustering more closely around neutral.

However, these baseline valences shift noticeably when the subjects are placed within sentences. For literal sentences, valence ratings tend to converge toward neutrality (around 5), regardless of the subject's inherent valence. This suggests that the literal use of manner of motion verbs focuses attention on the physical act of movement, dampening the influence of the subject's emotional tone.

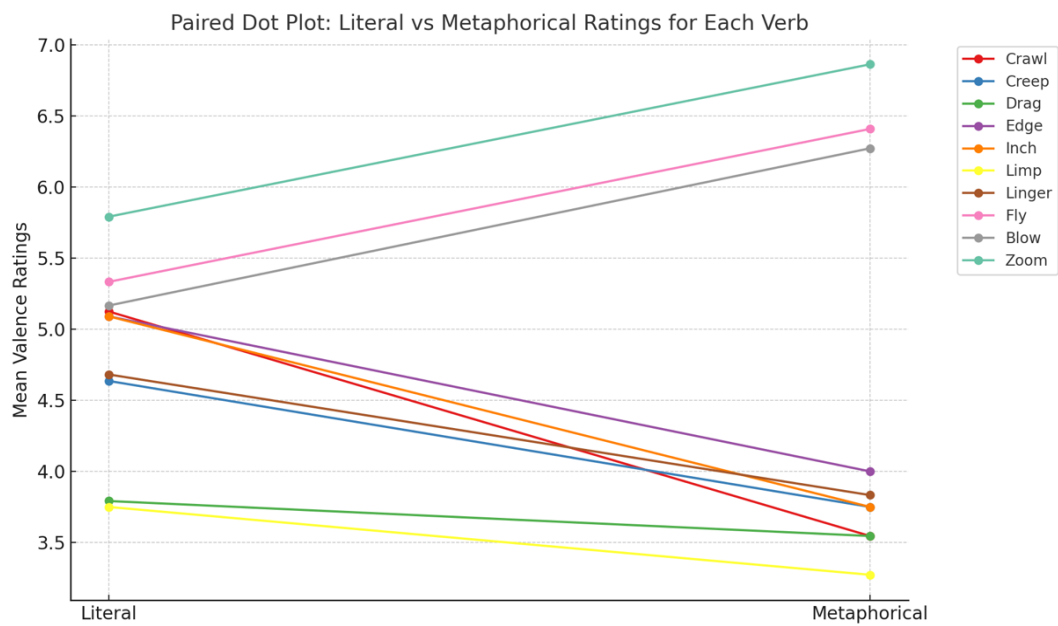

Figure 9. Pair dot plot: Literal vs metaphorical mean valence ratings for each verb

In contrast, temporal sentences exhibit greater variability in valence ratings, driven by the interaction between manner of motion verbs and temporal subjects. For instance, metaphorical expressions like *the week crept by* evoke a more negative emotional tone than *the ant crept by*, while *the hours flew by* elicit a more positive one than *the plane flew by*. These findings highlight how the metaphorical use of manner of motion verbs reshapes emotional interpretation, amplifying valence shifts even when paired with relatively neutral temporal subjects. The results provide evidence that manner of motion verbs elicit distinct emotional valences when used metaphorically to describe time.

**Conclusions**

This study has explored how manner of motion verbs contribute to emotional valence in literal and metaphorical contexts, providing insight into the dynamic process of meaning construction. In literal contexts, where these verbs describe physical motion, valence ratings tended to remain near neutrality, suggesting that their meaning was primarily shaped by the physical properties of the described movement. In metaphorical contexts, however, significant shifts in valence were observed. Slow verbs such as *crawl* and *drag* elicited more negative valence, often associated with tedious or undesirable time progression, whereas fast verbs like *fly* and *zoom* were linked to more positive valence, reflecting perceptions of dynamic or enjoyable experiences.

These findings suggest that meaning is constructed not only from the individual lexical items but also from their interaction with the conceptual domains they evoke. In literal contexts, the interaction is grounded in the physical domain, resulting in more stable valence ratings. In metaphorical contexts, however, the interaction between manner of motion verbs and temporal units appears to give rise to emergent emotional meanings that are not inherent to the temporal units themselves. This highlights how the combination of linguistic elements, such as verbs and subjects, can create novel meanings when extended to abstract domains like time.

## 2.3. Study 3: Sentence-emoji association

**Participants**

This study involved 32 native English speakers aged between 18 and 27; 24 were females and 9 were males. All participants were undergraduate students in the Department of Cognitive Science at the University of California, San Diego.

**Materials**

The materials for this experiment comprised 14 sentences and 11 emojis (see Appendix III). Each sentence incorporated a motion verb used with a TMU serving as the subject. The motion verbs included were *slip, creep, inch, blow, limp, drag, crawl, linger, fly*, and *rush*, due to their potential to activate complex meanings beyond their conventional physical uses when integrated within temporal contexts, as seen in previous corpus analyses (Valenzuela & Illán Castillo, 2022; Illán Castillo, 2024). Following the presentation of each sentence, participants were shown two emojis (out of the 11 emojis selected): one that was congruent with the expected emotional meaning of the verb in temporal contexts, and another that was incongruent. The selection of emojis for this experiment was based on a prior validation process to ensure that each emoji's emotional significance was unmistakably aligned with either positive or negative valence. During the validation phase, participants[1] evaluated whether each emoji conveyed positive, negative, or neutral sentiments. This preparatory step was crucial to guarantee that the emojis employed in the main study would effectively reflect the intended emotional dimensions.

---

[1] The participants in the validation study were distinct from those who engaged in the final experimental task.

**Procedure**

The study was conducted in person using the Qualtrics platform, allowing for a controlled environment and direct interaction with participants. This setup facilitated the accurate capture of participants' immediate emotional responses to the sentence-emoji pairings. In the task, participants were sequentially presented with sentences that featured motion verbs in temporal usage. After reading each sentence, the two pre-validated emojis —one congruent and one incongruent with the verb's emotional tone— were displayed. Participants were then tasked with selecting the emoji they felt best captured the emotion evoked by the preceding sentence. Figure 10 is an example:

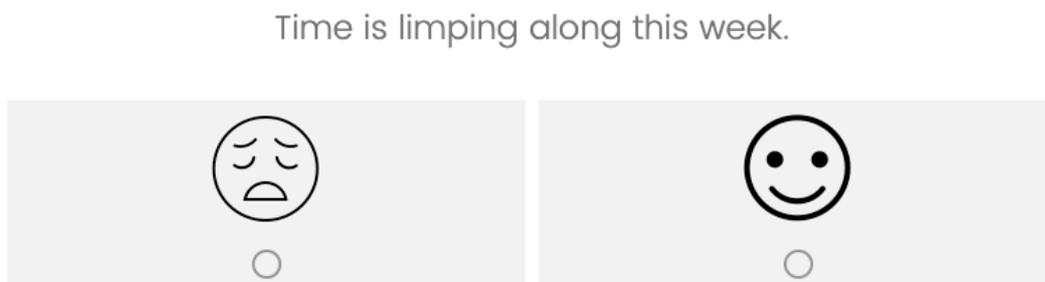

Figure 10. Question sample in the sentence-emoji association experimental task

**Results and discussion**

Results show a clear link between the manner of motion verbs and certain emotions:

Example result 1) *Time is limping along this week*.

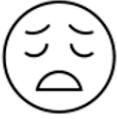

Figure 11. Example result for *Time is limping along this week*

Example result 2) *Time flew by during our dinner*.

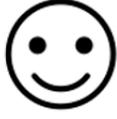

Figure 12. Example result for *Time flew by during our dinner*

We found an evident connection between manner of motion verbs and either positive or negative valence associated with them when used in temporal contexts:

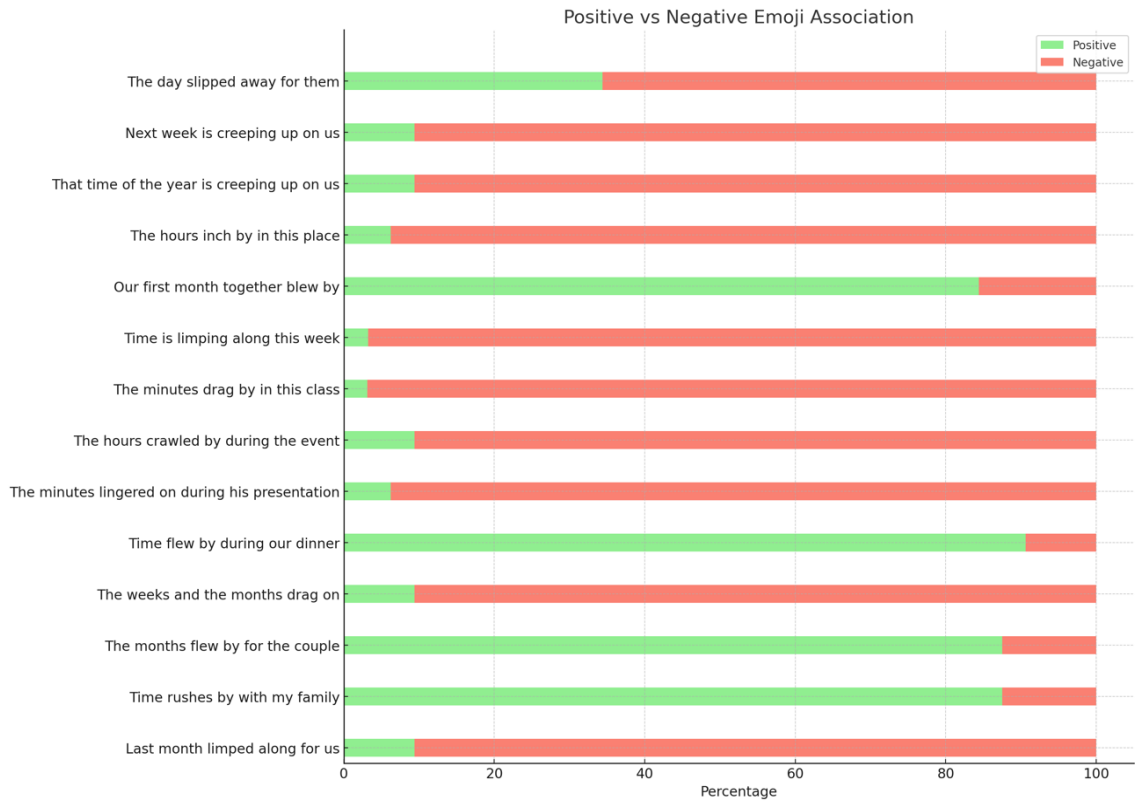

Figure 13. Positive - Negative emoji association

Our results show that most of the verbs are clearly associated with the two poles of the valence scale by our participants. The negative sentiment category includes the verbs *drag*, *linger*, *crawl*, *creep*, *limp*, and *inch*, which are typically associated with slow, tedious, or challenging progress. The analysis revealed that descriptions containing these verbs had an average positive sentiment of 8.04% and a negative sentiment of 91.96%, underscoring their strong negative sentiment associations. This suggests that the usage of these verbs in descriptions conveys a predominately negative experience or perception when they are used in temporal contexts.

The positive sentiment category comprises the verbs *fly*, *blow* and *rush*. This category captured descriptions that convey quick, efficient movement or the passing of time, often with a sense of achievement or excitement. The statistics for this category showed an average positive sentiment of 88.54% and a negative sentiment of 11.46%, indicating that these verbs are strongly associated with positive experiences or perceptions when used in temporal contexts.

Outside of these two main categories, we find the verb *slip*, with an average positive sentiment of 32.06% and a negative sentiment of 67.94%. When used in the context of time, this verb embodies a duality in sentiment that makes its classification less straightforward compared to verbs that clearly connote positive or negative sentiments. This ambiguity can be attributed to the nuanced implications *slip* carries, particularly in relation to the passage of time. On one hand, this verb can convey a sense of quick passage of time, akin to the way *fly*, *blow* and *rush* are perceived. This quickness often aligns with enjoyment or positive experiences —as seen in the previous two experiments— where time *slipping away* suggests engaging activities or moments that are so captivating that one loses track of time. This association could place *slip* closer to a positive sentiment, as it implies moments filled with pleasure or satisfaction that make time seem to pass more rapidly than usual. On the other hand, *slip* also carries a connotation of loss of control or lack of productivity. This aspect highlights an involuntary or unintentional passing of time, where moments *slip away* without one's consent or awareness. It can evoke feelings of regret or a sense of missed opportunity, as time has passed without being fully utilized or appreciated. This interpretation aligns more closely with a negative sentiment, emphasizing the uncontrollable and fleeting nature of time that escapes grasp.

Context would be then the element triggering if the verb's meaning that is activated in a sentence leans more towards one sense or another.

**Conclusions**

This study has elucidated the nuanced interplay between verb usage and sentiment perception in the context of time-related experiences. Through a systematic analysis of sentiment ratings associated with specific verbs, we have uncovered distinct patterns that underscore the profound impact of lexical choice on emotional resonance. The findings reveal a clear dichotomy between verbs that convey positive sentiments, such as *fly*, *blow* and *rush*, and those that evoke negative sentiments, including *drag*, *linger*, *crawl*, *creep*, and *inch*. Interestingly, the verb *slip* occupies a middle ground, reflecting the complexity of the meaning construction processes we are addressing and the subtleties of language in conveying emotions throughout the use of motion verbs in temporal contexts.

## 2.4. Study 4: Fill in the blanks with path and manner verbs

**Participants**

A total of 32 individuals participated in this experiment. The participant pool consisted of 24 females and 9 males. All participants were native speakers of English, ensuring linguistic proficiency for the task. They were all undergraduate students in the Department of Cognitive Science at the University of California, San Diego, and each provided informed consent prior to the study.

**Materials**

The experimental design incorporated a fill-in-the-blank task with 12 sentences (see Appendix IV). Each sentence contained a blank space where participants were instructed to select the motion verb that best suited the context of the sentence. The verbs presented for selection fell into two categories: manner verbs and path verbs. Participants were provided with a choice between two manner verbs in some sentences and with a choice between a manner and a path verb in others. Out of the 12 sentences, half included a choice between two manner verbs, and the other half offered a choice between a manner and a path verb. All participants completed the task with the full set of sentences, which included both types of verb choices. The sentences selected as stimuli had explicitly included emotions (e.g. *His speech was monotonous. Those two hours…(dragged/passed) by*; *We had a wonderful time at the beach. It felt like the hours…(crawled/rushed) by*.

**Procedure**

Participants were individually seated in a quiet room and presented with the sentences on a computer screen, through Qualtrics, one at a time. They were instructed to read each sentence and select the verb that they felt best completed it. The sentences were randomized to control for order effects, and participants were allowed to take as much time as needed to make their selection, ensuring thoughtful responses.

**Results and discussion**

The results were highly indicative of a congruence between the verbs chosen and the emotional content of the sentence. When participants were presented with the option to choose between two manner verbs, their selections consistently aligned with the emotional content explicitly expressed in the sentences. For example, given a sentence that depicted a tedious event, the verb *limping*, which suggests a slow passage of time — as opposed to *flying*—, was preferred overwhelmingly. Similarly, for sentences conveying enjoyable events, verbs that suggest rapid passage, such as *rush* and *zoom*, were favored almost unanimously. This pattern indicates that participants clearly relate the meanings conveyed by manner of motion verbs in temporal contexts to some specific emotions and not others.

In contrast, when participants were provided a choice between a manner verb and a path verb, they invariably opted for the manner verb. This suggests a distinct preference for manner verbs in the presence of emotional content. The decision to choose manner verbs, which can activate connotative emotional weight, over more neutral path verbs, may highlight the cognitive inclination to emphasize the emotional aspects of an experience when describing the passage of time. Below is a summary of the results obtained.

In the selection of motion verbs, a remarkable preference was observed for verbs associated with certain subjective perceptions of time that were linked to the emotions explicitly stated in the sentence to be completed. Specifically, when regarding a sentence where the explicit emotion encoded was *tedious*, a highly significant majority of 93.5% opted for *limping along* as opposed to *flying by*, yielding χ^2(1, N=32) = 49.0, p < .001, which indicates a strong preference for a verb that corresponds with the negative emotional valence of *tedium*. Conversely, in scenarios denoting positive experiences, such as leisure time at the beach, an overwhelming 96.88% of choices favored *rushed by* over *crawled by*, resulting in a χ^2(1, N=32) = 56.25, p < .001. This, once again, illustrates a pronounced alignment of verb selection with the positive emotional tone of the sentence. Moreover, in the context of choosing between a manner and a path verb, when participants were presented with a description where the focus was on productivity or lack thereof, the choice of verb also significantly leaned towards those imparting an emotional undertone. For instance, in reflecting on a non-productive day, the manner verb *slipped away* was chosen by 87.5% over the path verb *went by*, with a χ^2(1, N=32) = 36.0, p < .001, underscoring a predilection for verbs that resonate with the sentiment of inefficiency and the fleeting nature of time. Additionally, in the depiction of a monotonous speech, 93.75% selected *dragged by* against *passed by*, computed at a χ^2(1, N=32) = 49.0, p < .001. The following graph presents a visual comparison of verb choices across five distinct contexts, highlighting the participants' preference for verbs that align with the emotional valence of each scenario:

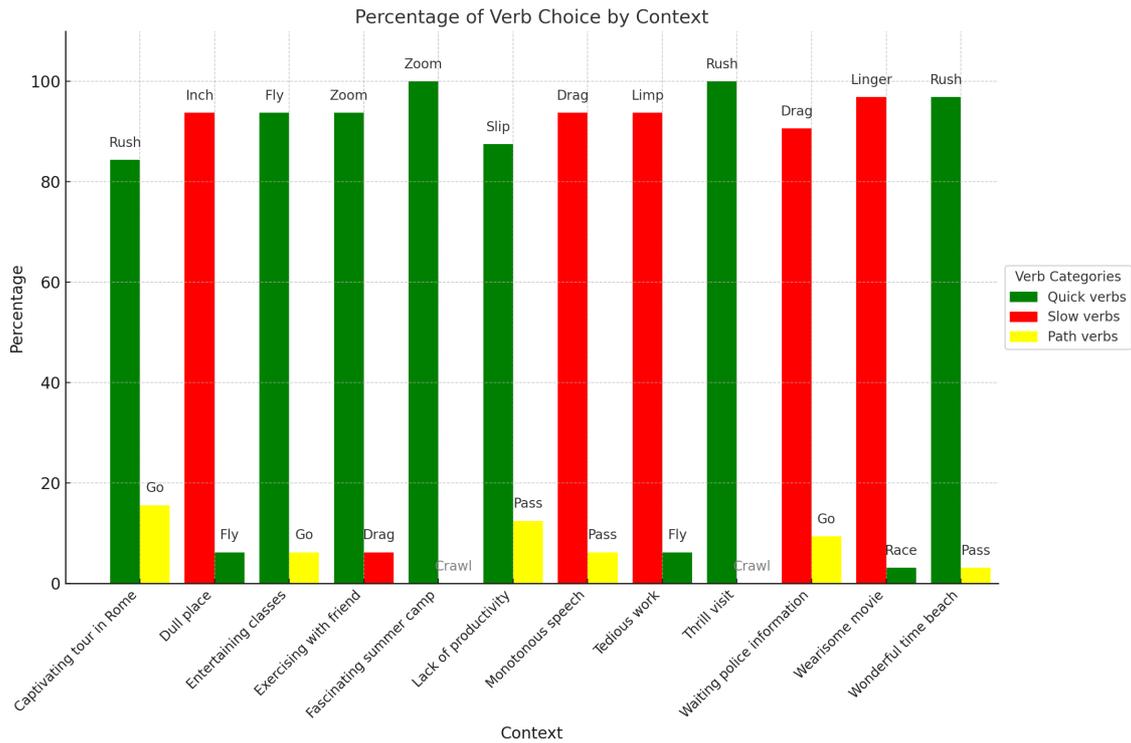

Figure 14. Percentage of verb choice by context

These unequivocal patterns underline the significance of motion verbs in temporal contexts, revealing their capacity to convey subtle nuances of emotional experience. As we saw in the previous studies, motion verbs play a crucial role in language by encoding not just physical movement but also temporal dynamics and emotional valence. Their selection in language use can significantly influence the listener's or reader's perception, imbuing the described events with a particular emotion or mood.

**Conclusions**

This study highlights the intricate relationship between motion verb selection, emotional valence perception, and contextual variation. By analyzing participants' verb choices across various scenarios, the study illustrates how the context in which a verb is used significantly shapes lexical selection and the interpretation of emotional experiences. The findings reveal that speakers adapt their use of motion verbs based on the nuances of the temporal and emotional context, demonstrating a preference for manner verbs over path verbs when expressing subjective experiences of time's passage. We found a strong association between manner verbs and emotional expression, as these verbs inherently convey nuances that align with subjective experiences of time's passage. In contrast, path verbs lack such an association, serving instead to describe the directional progression of time without encoding specific emotional tones. In contexts where emotional resonance is pivotal and noticeably present, speakers clearly opt for using manner verbs, which allow them to encode and communicate these subtleties more effectively. This adaptability highlights the dynamic interaction between lexical choice and context, showing how language users leverage the expressive potential of manner verbs to align their message with the situational demands of the moment.

## 3. Conclusions

The research presented in this paper provides a multifaceted exploration of how manner-of-motion verbs shape subjective temporal perception, emotional resonance and meaning construction. Across four studies we demonstrate that these verbs play a significant role in influencing and encoding how time is experienced, beyond simply indicating its passage. Manner verbs significantly impact how individuals perceive the flow of time, shaping not only its pacing but also adding a number of emotional nuances.

We observed a strong positive correlation between speed and perceptions of liveliness, eventfulness, and control, with faster motion verbs evoking a more dynamic and engaging temporal experience. Faster verbs were associated with greater perceived agency, likely because they imply a heightened level of involvement in the experience. In contrast, slower verbs were linked to a more passive, monotonous, and less controllable experience of time. This nuanced distinction suggests that the implied speed of motion verbs acts as a cognitive cue, influencing whether time is felt as actively engaging or passively endured.

Additionally, we found a clear association between manner of motion verbs and emotional valence. Faster verbs were generally associated with positive emotional experiences, while slower verbs tended to align with negative emotions. This finding suggests that manner verbs not only convey temporal speed but also carry emotional weight when associated with time, further shaping the subjective quality of temporal experiences. While literal uses of motion verbs tend to remain emotionally neutral, metaphorical-temporal uses generate significant shifts in valence, with slower verbs amplifying negative emotions and faster verbs reinforcing positive ones. These shifts demonstrate how language blends features of motion, emotion, and time, creating nuanced meanings. The interplay of manner features and time reinforces how these verbs shape temporal perception and conceptualization by adding layers of experiential and emotional nuance. The fill-in-the-blank task supports a clear preference for specific manner verbs that align with the emotional tone encoded in each sentence. When an emotion is explicitly associated with a temporal event, participants consistently chose manner verbs (e.g. *drag, fly*) over path verbs (e.g. *pass, go*) to describe the passage of time, favoring verbs that match the emotional and control dimensions of the experience.

These insights highlight the role of linguistic framing in structuring temporal perception and emotional interpretation. Manner verbs do more than describe time —they actively shape how it is felt and understood. These verbs enrich temporal language and reveal the intricate connections between motion, cognition, and experience.

Overall, this paper advances our understanding of how manner-of-motion verbs contribute to the conceptualization of time in language. These findings support the notion that temporal perception is shaped not only by spatial metaphors but also by linguistic choices that blend aspects of motion and emotion. Future research could explore these dynamics across different languages and contexts to further clarify the intricate relationship between motion and temporal perception.

APPENDIX I

**Stimuli for study 1: Emergent meanings survey (English)**

- The new year is approaching.
- The two hours that the show lasted flew by.
- The minutes hasten in this place.
- Our dinner together raced by.
- The days rush when I'm home.
- Last week slipped by.
- The days slid into January.
- We will know more as the day spins on.
- Election day is inching closer.
- The first full day of trial dragged on.
- The days creep towards the end of the week.
- The hours crawled by that afternoon.
- This month is limping along.
- The hour edged closer.
- The news hour runs from 5pm to 7pm.

APPENDIX II

**Stimuli for study 2: Constructing meaning and valence - Manner of motion verbs in literal (physical) and metaphorical (temporal) contexts**

1. **Crawl**
   - Literal: The ant crawled by.
   - Temporal: The hours crawled by.
2. **Creep**
   - Literal: The ivy crept up.
   - Temporal: The minutes crept by.
3. **Drag**
   - Literal: He dragged it along.
   - Temporal: The days dragged along.
4. **Edge**
   - Literal: The turtle edged along.
   - Temporal: The weeks edged along.
5. **Fly**
   - Literal: The plane flew by.
   - Temporal: The years flew by.
6. **Inch**
   - Literal: The snail inched by.
   - Temporal: The hours inched by.
7. **Limp**
   - Literal: He limped forward.
   - Temporal: The days limped forward.
8. **Linger**
   - Literal: His finger lingered on the glass.
   - Temporal: The minutes lingered on.
9. **Zoom**
   - Literal: The motorcycle zoomed by.
   - Temporal: The minutes zoomed by.
10. **Blow**
    - The wind blew by
    - The hours blew by

APPENDIX III

**Stimuli for study 3: Sentence-emoji association**

1) **Sentences:**

- The day slipped away for them.
- Next week is creeping up on us.
- That time of the year is creeping up on us.
- The hours inch by in this place.
- Our first month together blew by.
- Time is limping along this week.
- The minutes drag by in this class.
- The hours crawled by during the event.
- The minutes lingered on during his presentation.
- Time flew by during our dinner.
- The weeks and the months drag on.
- The months flew by for the couple.
- Time rushes by with my family.
- Last month limped along for us.

2) **Emojis:**

  o Positive valence:

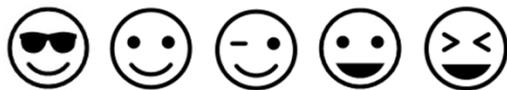

  o Negative valence:

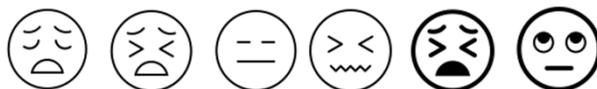

APPENDIX IV

**Stimuli for study 4: Fill in the blanks with path and manner verbs**

- Exercising with a friend can make the minutes ______ by. (Zoom / Drag)
- Our work last week was tedious. It felt like the days were ______. (Flying by / Limping along)
- We were thrilled with their visit. During our time together the hours ______ by. (Rushed / Crawled)
- The summer camp was fascinating. Three weeks ______ by. (Crawled / Zoomed)
- This place is dull. The hours ______ by. (Inch / Fly)
- For next week's class we must write about a wearisome movie. When I tried to watch it the hours ______. (Lingered on / Raced by)
- While she waited for the police to give her some information about her missing brother, the minutes ______ by. (Dragged / Went)
- Our tour around Rome was captivating. The three days we spent there ______ by. (Rushed / Went)
- It's been just two days and I already love this teacher. During her classes time ______ by. (Goes / Flies).
- His speech was monotonous. Those two hours ______ by. (Dragged / Passed)
- We were not very productive yesterday. Our day ______. (Slipped away / Passed by by)
- We had a wonderful time at the beach. The day ______. (Rushed by / Passed by).